\documentclass[cameraready]{Interspeech}

\title{Learning to Hear Hesitation: Continual Learning for Disfluency-Aware ASR}

\author[orcid=0009-0001-0691-7819, equalcontribution]{Henri-Leon}{Kordt}
\author[orcid=0000-0002-2847-5097, equalcontribution]{Theresa}{Pekarek Rosin}
\author[orcid=0000-0001-9840-780X]{Jae Hee}{Lee}
\author[orcid=0000-0003-1343-4775]{Stefan}{Wermter}

\address{
    Knowledge Technology, Department of Informatics, University of Hamburg, 
    Germany
}

\email{\{henri-leon.kordt, theresa.pekarek-rosin, jae.hee.lee, stefan.wermter\}@uni-hamburg.de}


\DeclareRobustCommand{\tblmethod}[1]{\textsc{#1}}
\DeclareRobustCommand{\tblmetric}[1]{\textbf{#1}}
\DeclareRobustCommand{\tblscore}[1]{\footnotesize\textrm{#1}}
\DeclareRobustCommand{\tbldataset}[1]{\textsf{#1}}
\usepackage{comment}
\usepackage{placeins}

\usepackage[babel=true,expansion=true,kerning=true,tracking=true,spacing=true,verbose=silent]{microtype}
\usepackage{inconsolata} 
\usepackage[capitalize]{cleveref}
\crefname{figure}{Figure}{Figures}

\newcommand{\pms}[1]{\textsuperscript{\tiny$\pm$#1}}
\begin{document}

\maketitle

\begin{abstract}
	Despite advances in large-scale Automatic Speech Recognition (ASR), disfluent speech remains challenging, as state-of-the-art systems are often optimized to omit disfluencies, leading to information loss and hallucinations. Prior work has focused on verbatim transcription and the integration of disfluency markers, but adapting models on limited datasets can lead to catastrophic forgetting of general-domain knowledge. We address this gap by leveraging continual learning (CL) with explicit disfluency tokens. We first introduce these tokens into a pretrained ASR model to establish stable token mechanisms, and then continue training on additional datasets with varying disfluency distributions. Through a detailed analysis of model dynamics during training, we identify a trade-off between marker learning and ASR performance, and a consistent cross-attention head mechanism shared across CL methods.  
\end{abstract}

\section{Introduction}

Although Automatic Speech Recognition has improved substantially in recent years due to the introduction of large-scale models trained on vast amounts of audio data~\cite{Radford2022}, disfluent speech recognition has not seen the same progress~\cite{FENG2024101567}. Disfluencies are perturbations in speech that can range in severity from simple filler words (``uh", ``uhm"), hesitations, and word or phrase repairs to slurred speech or stutters caused by underlying conditions.
However, state-of-the-art models are often trained to ignore disfluencies in order to produce clean transcriptions~\cite{hwang-etal-2025-evaluating,mujtaba-etal-2024-lost}, which can lead to incorrectly generated utterances, i.e., hallucinations or fabrications, and to a general loss of information and transparency~\cite{lin25b_slate,PGKKW25}.

To address this problem and improve performance on disfluent speech, MacDonald et al.~\cite{MacDonald2021euphonia} have suggested personalizing models for each speaker, while others have focused on creating more verbatim transcriptions~\cite{lin25b_slate,zusag24_interspeech}. Such verbatim transcriptions and the subsequent inclusion of disfluencies can capture clinically relevant speech information~\cite{soleimani2024}, improve audio-text alignment for timestamping~\cite{zusag24_interspeech}, and even improve downstream ASR performance~\cite{lin25b_slate,whisperd}.
Previous work has also introduced specific markers or tokens to represent speech disfluencies by training pretrained models on additional annotated speech corpora, both to detect dementia from utterances~\cite{soleimani2024} and to improve performance on disfluent~\cite{whisperd} or spontaneous speech~\cite{zusag24_interspeech}.
\begin{figure}[ht]
	\centering
	\includegraphics[width=\linewidth]{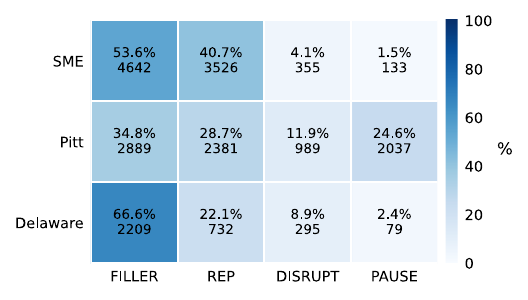}
		\caption{The distribution of disfluencies across the selected datasets. Percentages are calculated with respect to each dataset's distribution.}
	\label{fig:disfluencydistributions}
\end{figure}
However, when applied carelessly to small-scale datasets, these approaches often lead to a loss of domain-agnostic performance~\cite{cl_survey}, an effect known as catastrophic forgetting~\cite{vandeven2024clcf}. A naive approach to address this effect is to jointly train the model on the new data and the entire pretraining data. Since speech patterns change over time and new data becomes available constantly, this process would have to be repeated for every new data instance, which is highly inefficient. Additionally, because training data for disfluent speech is often limited and subject to privacy regulations, this approach is frequently infeasible.

Continual Learning (CL) is a paradigm for mitigating catastrophic forgetting by balancing \textit{learning plasticity} and \textit{memory stability} through regularization, replay, or architectural approaches~\cite{cl_survey}. Such methods provide an efficient way to integrate heterogeneous datasets and speech patterns without sacrificing the pretrained model's performance. CL has been successfully applied to domain adaptation for non-standard speech~\cite{vandereeckt23_interspeech,PW23} and low-resource languages~\cite{dellalibera_clmasr,song24_interspeech}, but it has not yet been utilized for disfluent or impaired speech.
Moreover, disfluency events and disfluency transcriptions pose unique challenges that differ from typical ASR domain adaptation tasks, for which evidence of CL effectiveness remains unexplored.

In this work, we address disfluent speech recognition using disfluency markers, which are commonly employed in disfluency-focused ASR research~\cite{lin25b_slate,zusag24_interspeech,whisperd} and are also present in the manual transcriptions of datasets collected under the TalkBank~\cite{talkbank_sla,dementiabank} umbrella.
We aggregate these markers into four token types and introduce them to a pretrained model using different CL methods to preserve the model's performance. We then continue training on two annotated datasets with different disfluency distributions to examine the robustness of the established marker representations for continual adaptation.
We examine adaptation differences across CL methods and find that, despite performance differences, successful marker learning largely follows a shared attention-head specialization pattern.

\section{Methodology}
To enable more verbatim transcriptions and increase the amount of information produced by the model, we introduce disfluency markers into a pretrained backbone model. We combine continual learning (CL) with explainability methods across three datasets to investigate internal specialization mechanisms while preserving performance under continual model adaptation.

\subsection{Datasets}

To cover both healthy and impaired disfluent speech, we use three datasets from the TalkBank repository~\cite{talkbank_sla, dementiabank}, which were selected based on data availability, number of samples, disfluency distribution, and domain: the Standard Malaysian English (SME) Corpus~\cite{sme_thesis}, the Pitt Corpus~\cite{pitt}, and the Delaware Corpus~\cite{dementiabank}. All three use the unified CHAT Transcription Format~\cite{chat}, which ensures consistency in the labeling syntax. Their transcriptions also contain manually annotated disfluency markers, allowing a direct mapping to disfluency tokens during training.

The SME Corpus~\cite{sme_thesis} contains \SI{11.79}{\hour} of English speech from Malaysian university students who are second-language learners. This dataset represents a common subgroup of speakers who are generally healthy but may exhibit disfluencies in their speech production due to the language gap. The Pitt Corpus~\cite{pitt} is a collection of \SI{21.30}{\hour} of speech from elderly speakers with dementia or Alzheimer's disease, while the Delaware Corpus~\cite{dementiabank} contains \SI{9.72}{\hour} of speech from people with mild cognitive impairment. Both Pitt and Delaware include speech from healthy individuals in their control groups, which we divide evenly across the training and validation splits. We include only \SI{12.11}{\hour} of the Pitt corpus in the CL setup to reduce dataset-size effects. We use these datasets to demonstrate the robustness of our approach across different severities and distributions of disfluencies (see \cref{fig:disfluencydistributions}). In the CL setup, we also include a small subset of LibriSpeech data~\cite{librispeech} for the rehearsal buffer in replay-based methods and as a representation of unperturbed speech.

\subsection{Continual learning setup} \label{cl}

We use \texttt{whisper-small.en}\footnote{\url{https://huggingface.co/openai/whisper-small.en}}, a pretrained monolingual model, as the backbone for the CL setup. To preserve the pretrained performance of the model and handle the different disfluency distributions in the datasets, we use four CL methods that are commonly employed for domain-incremental ASR training: Elastic Weight Consolidation (EWC)~\cite{ewc}, Experience Replay (ER)~\cite{experiencereplay}, A-GEM~\cite{AGEM}, and Weight Averaging (WA)~\cite{vandereeckt23_interspeech}.

EWC~\cite{ewc} is a regularization-based method that identifies model parameters important to previously learned tasks and penalizes updates to those parameters (using the diagonal Fisher information matrix). For ER~\cite{experiencereplay}, a small percentage of the old data is kept in memory (rehearsal buffer) and used to supplement the training data for the new task, preventing drastic gradient updates. A-GEM~\cite{AGEM} is a gradient-based method that calculates the dot product between the gradients of the rehearsal buffer and the training data of the current dataset to penalize gradients in opposing directions. For WA~\cite{vandereeckt23_interspeech}, the weights of the old model are kept in memory, while another version is trained on the new task. Once training is complete, the model is updated with an average of the old and the new weights.

To evaluate general ASR performance, we use the preprocessed WER (pWER), in which punctuation, special characters, and disfluency markers are removed from the transcripts before the metric is computed.
To assess the model's memory stability and learning plasticity during CL, we use the metrics defined by Wang et al.~\cite{cl_survey}, with minor modifications to accommodate pWER and marker F1 in place of accuracy.
Average Word Error Rate (A-WER) and Average Incremental WER (AI-WER) measure overall pWER performance at the current step and across the full CL trajectory, respectively. For marker prediction, we similarly report Average F1 (A-F1) and Average Incremental F1 (AI-F1), which replace per-task accuracy with per-task marker F1. The F1 scores are calculated using a bag-of-words approach.
Backward Transfer (BWT) and Forgetting Measure (FM) assess memory stability, whereas Forward Transfer (FWT) and Intransigence Measure (IM) assess the model's learning plasticity.

\subsection{Explainability methods}
\label{xai}

Given the tendency of transformer models to develop specialization and redundancy~\cite{NEURIPS2019_2c601ad9,ahrens_visually_2023}, we investigate whether the transcription of disfluency events induces specialized circuits within the model. This is particularly relevant in the CL setting, since CL methods may yield similar aggregate error rates while inducing different adaptation dynamics and internal strategies. By probing attention heads, we aim to assess whether disfluency handling concentrates in a small set of heads (specialized circuitry) and how stable these mechanisms are across CL approaches, thereby providing insight into robustness and potential method-specific trade-offs. 
We follow the head-masking approach of Michel et al.~\cite{NEURIPS2019_2c601ad9}, which introduces a learnable scalar gate $\xi_h$ per attention head $h$. To estimate token-level head importance $I_h$, we compute the sensitivity of a token objective to $\xi_h$ and average over a set of token instances $X_t$:
\begin{align}
	I_h = \mathbb{E}_{x \sim X_t}\left[\frac{\partial \mathcal{L}(x)}{\partial \xi_h}\right]
\end{align}
Based on this per-head importance $I_h$, we rank head attribution for disfluency tokens and compare importance rankings averaged over non-disfluency tokens. Specifically, we calculate the top-10 lift, defined as the difference in the frequency with which a head $h$ appears in the top-10 ranking for disfluency tokens versus its frequency in the baseline distribution (all other tokens):
\begin{align}
	\text{Lift}_h = P(h \in \text{Top-}k \mid X_t) - P(h \in \text{Top-}k \mid X_{\text{base}})
\end{align}
To confirm head attribution, we ablate heads that show high top-10 lift scores using zero-masking. We inspect changes in token emission and control for overall performance using pWER.

\subsection{Experiments}
To assess the introduction of disfluency tokens into a pretrained model and the resulting task performance in a Continual Learning (CL) setup, we conduct two experiments: (i) disfluency token introduction and (ii) sequential continual adaptation. Following the literature~\cite{zusag24_interspeech,whisperd}, we aggregate similar disfluency events into four distinct token types: filler words (FILLER; e.g., ``um", ``uh"), repetitions/revisions (REP; word- and phoneme-level), disruptions (DISRUPT; e.g., coughs, laughs), and pauses (PAUSE).

In the \textbf{disfluency token introduction} experiment (cf.~\cref{sec:disfluencytokenintro}), we focus on introducing these tokens into a pretrained model without overfitting to the new dataset, i.e., while keeping performance on non-disfluent speech high. We fine-tune the backbone on the SME dataset with disfluency tokens using the CL techniques discussed above. SME covers the domain closest to the backbone and predominantly contains only two marker types: FILLER and REP. Additionally, we analyze whether the introduction of disfluency tokens is associated with the emergence of internal model mechanisms and whether these mechanisms differ across CL methods.

For the \textbf{sequential continual adaptation} experiment (cf.~\cref{sec:sequentialcl}), we choose the best-performing model from the disfluency token introduction stage: a model that shows successful marker integration (high marker-F1) while maintaining the smallest pWER trade-off. We then continue expanding the model with the Pitt and Delaware corpora, two disfluent-speech datasets that vary in domain and disfluency distribution (see Figure~\ref{fig:disfluencydistributions}). In doing so, we approximate lifelong model adaptation to changing domains and speaker variations through sequential training on more difficult datasets.
We begin with the Pitt Corpus, which introduces a substantial number of pause markers and also increases the number of disruptions. Afterward, we train the model on the Delaware Corpus, which contains the fewest markers overall and therefore specifically challenges the model to retain them.

We ensure that the training and evaluation splits for both experiments are speaker-disjoint and partition each dataset into an 80/20 split. For the rehearsal buffer, we randomly sample 10\% of the training data from each dataset, greedily prioritizing rare disfluency markers so that the token distribution mirrors that of the full dataset. We train on each dataset for 10 epochs with a learning rate of 2e-5 and a batch size of 16, ensuring convergence in both marker scores and ASR performance. Results are averaged across three runs. Following the literature~\cite{ewc,experiencereplay} and our empirical evaluation, we include 25\% of old data per batch, sampled from the rehearsal buffer for ER, and set the importance parameter to 1000 for EWC. In addition to the general performance and CL metrics discussed in Section~\ref{cl}, we also measure disfluency-marker recognition using micro- and macro-averaged F1 scores. For the disfluency token introduction experiment, we use micro-F1 to summarize overall marker integration for the introduced markers, while macro-F1 is used later to assess per-type balance once all markers are in scope.

\section{Results}
\subsection{Disfluency token introduction}
\label{sec:disfluencytokenintro}
\begin{table}[t]
	\centering
	\caption{Disfluency token introduction: The preprocessed Word Error Rate (pWER) averaged over 3 seeds after 10 epochs of training for SME and LibriSpeech (LS). We additionally report the micro-F1 score for marker predictions on SME.}
	\setlength{\tabcolsep}{.3em}
	\renewcommand{\arraystretch}{1.05}
	\begin{tabular}{l c c c}
		\toprule
		                     & \tbldataset{SME} (\tblscore{pWER}\% $\downarrow$) & \tbldataset{LS} (\tblscore{pWER}\% $\downarrow$) & \tbldataset{SME} (\tblscore{F1} $\uparrow$) \\
		\midrule
		\tblmethod{Backbone} & $15.97 \phantom{\pms{0.08}}$                      & $3.47 \phantom{\pms{0.08}}$                       & $0.00 \phantom{\pms{0.08}}$                 \\
		\midrule
		\tblmethod{FT}       & $12.21 \pms{0.18}$                                & $5.06 \pms{0.02}$                                 & $0.73 \pms{0.01}$                           \\
		\midrule
		\tblmethod{A-GEM}    & $12.17 \pms{0.17}$                                & $4.39 \pms{0.07}$                                 & $\textbf{0.75} \pms{0.01}$                           \\
		\tblmethod{ER}       & $12.47 \pms{0.03}$                                & $4.55 \pms{0.13}$                                 & $0.73 \pms{0.01}$                           \\
		\tblmethod{EWC}      & $10.34 \pms{0.47}$                                & $4.43 \pms{0.07}$                                 & $0.21 \pms{0.07}$                           \\
		\tblmethod{WA}       & $\phantom{0}\textbf{9.64} \pms{0.08}$                      & $\textbf{3.41} \pms{0.01}$                                 & $0.00 \pms{0.00}$                           \\
		\bottomrule
		\label{exp1_sme_sl}
	\end{tabular}

\end{table}

We train on SME using different CL methods to examine how well performance on non-disfluent speech (LibriSpeech test-clean, LS) can be retained while improving performance on the new domain and introducing disfluency markers into the model. \cref{exp1_sme_sl} shows that all CL methods retain LS performance substantially better than the FT baseline while also improving pWER on SME. Weight Averaging (WA) outperforms the other methods by retaining, and even slightly improving, the ASR model performance on LS while achieving the lowest pWER on SME at 9.64\%. EWC also improves pWER on SME relative to FT, but shows weaker LibriSpeech pWER retention than WA. Importantly, both methods fail to integrate markers reliably. The FT baseline, ER, and A-GEM consistently produce marker micro-F1 scores between 0.73 and 0.75, whereas EWC remains at a micro-F1 of 0.21 and WA fails to emit markers at all. This means that although A-GEM, ER, and the FT baseline perform worse on the trained domain, improving the pWER only to around 12\%, they successfully produce disfluency markers.

\begin{figure}[t]
	\centering
	\includegraphics[width=\linewidth]{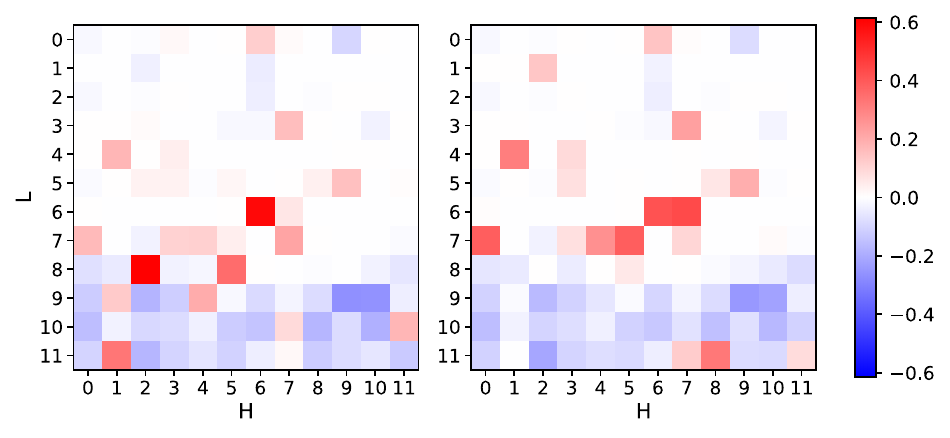}
		\caption{Top-10 lift scores for FILLER (left) and REP (right) across decoder cross-attention heads computed on the train split averaged across all token emitting methods.} 
	\label{fig:top10lift_raw_51870_mean}
\end{figure}

These results indicate a trade-off between pWER and marker performance, suggesting that marker training may require parameter updates that conflict with low-cost general ASR domain adaptation.
To test whether marker learning corresponds to a consistent mechanism that differs from normal token production, we apply our head-attribution approach (Section~\ref{xai}) across all methods. We find that when markers are successfully trained, a few specialized decoder cross-attention heads are consistently assigned to marker emission, as shown in \cref{fig:top10lift_raw_51870_mean}.

Notably, the same small set of heads emerges across all marker-emitting methods, suggesting a shared cross-attention mechanism rather than method-specific marker strategies induced by different regularization. In \cref{tab:ablation}, we verify their causal role through an ablation study on the top-5 cross-attention heads associated with disfluency markers and show that we can consistently remove around 50\% of the emitted FILLER markers while retaining similar pWER values.
\begin{table}[t]
	\centering
		\caption{Zero-mask ablation, averaged across methods. We report the relative change in marker emission (\%) and the change in pWER after masking the top-5 cross-attention heads identified for FILLER, REP, their union, and a random-head control.}
	\renewcommand{\arraystretch}{1.05}
    
	\begin{tabular}{l r r r}
		\toprule
		                      & $\Delta$\tblscore{FILLER} & $\Delta$\tblscore{REP} & $\Delta$\tblscore{pWER\%}  \\
		\midrule
		FILLER: Top-5         & $-57.0 \pms{5.2}$         & $-7.1 \pms{3.8}$       & $+0.42 \pms{0.64}$      \\
		REP: Top-5            & $-2.8 \pms{1.1}$          & $-15.7 \pms{4.2}$      & $+0.13 \pms{0.46}$      \\
		$\bigcup$ Top-5       & $-62.2 \pms{5.2}$         & $-21.2 \pms{9.8}$      & $+0.44 \pms{0.68}$      \\
		Control ($5\times r$) & $+0.2 \pms{0.5}$          & $+0.4 \pms{1.2}$       & $+0.15 \pms{0.21}$      \\
		\bottomrule
	\end{tabular}
	\label{tab:ablation}
\end{table}

\subsection{Sequential continual adaptation}
\label{sec:sequentialcl}

\begin{table*}[t!]
	\centering
	\caption{Continual learning metrics after sequential adaptation on Pitt and Delaware, initialized from the selected disfluency token introduction checkpoint. For each metric, we report pWER-based and marker F1-based variants.}
	\label{tab:big-cl}
	\footnotesize
	\setlength{\tabcolsep}{3pt}
	\renewcommand{\arraystretch}{1.15}
	\begin{tabular}{l cc cc cc cc cc cc}
		\toprule
		                                    & \multicolumn{2}{c}{\tblmetric{A-WER\%} / \tblmetric{A-F1}} & \multicolumn{2}{c}{\tblmetric{AI-WER\%}/\tblmetric{AI-F1}} &
		\multicolumn{2}{c}{\tblmetric{BWT}} & \multicolumn{2}{c}{\tblmetric{FM}}                    &
		\multicolumn{2}{c}{\tblmetric{FWT}} & \multicolumn{2}{c}{\tblmetric{IM}}                                                                                                                                                                                                                                                                                                                                   \\
		\cmidrule(lr){2-3}\cmidrule(lr){4-5}\cmidrule(lr){6-7}\cmidrule(lr){8-9}\cmidrule(lr){10-11}\cmidrule(lr){12-13}

		                                    & \tblscore{pWER} $\downarrow$                          & \tblscore{F1} $\uparrow$
		                                    & \tblscore{pWER} $\downarrow$                          & \tblscore{F1} $\uparrow$
		                                    & \tblscore{pWER} $\uparrow$                            & \tblscore{F1} $\uparrow$
		                                    & \tblscore{pWER} $\downarrow$                          & \tblscore{F1}  $\downarrow$
		                                    & \tblscore{pWER} $\uparrow$                            & \tblscore{F1} $\uparrow$
		                                    & \tblscore{pWER} $\downarrow$                          & \tblscore{F1} $\downarrow$                                                                                                                                                                                                                                                                                   \\
		\midrule
		\tblmethod{JOINT}                   & 17.95\pms{0.28}                                           & 0.47\pms{0.01}                                                    & --                       & --                      & --                       & --                      & --                      & --                      & --                       & --              & --                      & --              \\
		\tblmethod{FT}                      & 20.24\pms{0.14}                                       & 0.39\pms{0.02}                                        & 19.00\pms{0.06}          & 0.49\pms{0.01}          & -3.18\pms{0.26}          & -0.16\pms{0.02}         & 3.48\pms{0.35}          & 0.19\pms{0.02}          & -3.81\pms{0.10}          & -0.05\pms{0.01} & 0.44\pms{0.07}          & -0.03\pms{0.01} \\
		\midrule
		\tblmethod{A-GEM}                   & 20.15\pms{0.68}                                       & 0.36\pms{0.02}                                        & 18.88\pms{0.24}          & 0.48\pms{0.01}          & -3.46\pms{0.78}          & -0.19\pms{0.03}         & 3.50\pms{0.72}          & 0.21\pms{0.03}          & \textbf{-3.39}\pms{0.24} & -0.07\pms{0.01} & \textbf{0.16}\pms{0.16} & -0.02\pms{0.00} \\
		\tblmethod{ER}                      & 19.71\pms{0.23}                                       & \textbf{0.49}\pms{0.01}                               & 18.17\pms{0.10}          & \textbf{0.53}\pms{0.01} & -2.57\pms{0.48}          & \textbf{0.01}\pms{0.01} & 2.59\pms{0.46}          & \textbf{0.02}\pms{0.00} & -3.63\pms{0.52}          & -0.06\pms{0.00} & 0.32\pms{0.34}          & -0.03\pms{0.00} \\
		\tblmethod{EWC}                     & 19.01\pms{0.08}                                       & 0.44\pms{0.01}                                        & 18.28\pms{0.05}          & 0.52\pms{0.00}          & -1.40\pms{0.08}          & -0.09\pms{0.02}         & 2.18\pms{0.16}          & 0.13\pms{0.02}          & -3.75\pms{0.20}          & \textbf{-0.05}\pms{0.01} & 0.40\pms{0.13}          & \textbf{-0.03}\pms{0.01} \\
		\tblmethod{WA}                      & \textbf{18.90}\pms{0.19}                              & 0.46\pms{0.00}                                        & \textbf{17.57}\pms{0.04} & 0.51\pms{0.01}          & \textbf{-0.55}\pms{0.34} & -0.01\pms{0.00}         & \textbf{1.35}\pms{0.20} & 0.05\pms{0.01}          & -4.43\pms{0.13}          & -0.10\pms{0.00} & 0.85\pms{0.09}          & 0.00\pms{0.00}  \\
		\bottomrule
	\end{tabular}
\end{table*}

For the sequential continual adaptation experiment, we investigate the ability of CL methods to generalize and retain knowledge when the model is continuously trained on more difficult disfluency datasets after an initial marker set has been established. We initialize all methods from the same disfluency-token introduction checkpoint (A-GEM Seed 2), which reliably emits markers, selected by SME marker-F1 and SME pWER, and then sequentially train on Pitt and Delaware using each CL strategy.
In \cref{tab:big-cl}, we report common CL metrics (cf.~Section~\ref{cl}) after fine-tuning on Pitt and Delaware using different CL techniques. 

Our results show that all CL methods reduce pWER relative to the FT baseline, as reflected in lower A-WERs across all methods and all methods except A-GEM also improve marker F1.
Importantly, the best-performing methods differ for ASR performance and marker retention: WA achieves an A-WER of 18.90\%, while ER shows the best marker performance with a macro-F1 score of 0.49.

WA shows the strongest pWER reduction while also exhibiting the most stable pWER performance, as indicated by AI-WER. WA achieves these gains mainly through retention, exhibiting minimal forgetting and stable backward transfer, while showing comparatively limited plasticity in learning new tasks, as reflected by the high IM score.

ER, on the other hand, achieves the best performance for disfluency-marker F1. The method also shows strong generalization for markers, with nearly no forgetting as indicated by an FM score of 0.02, and even slight improvements on old tasks, reflected in a slightly positive marker-F1 BWT of 0.01.

To understand which marker types drive the overall F1 differences, we report the final marker-F1 scores for the different disfluency markers in \cref{tab:f1-exp2}. FILLER markers are robust across methods, with F1 scores between 0.68 and 0.75, whereas REP markers exhibit the strongest method sensitivity, ranging from 0.35 to 0.63, indicating learning differences or forgetting. PAUSE markers are the most challenging overall, but ER yields a clear advantage, which is consistent with ER achieving the best aggregate marker F1 in Table~\ref{tab:big-cl}.

We additionally evaluate pWER for LS to isolate non-disfluent speech retention after sequential training and find that the ranking closely matches the disfluent-speech retention trend: WA performs best by a clear margin (4.68\%), EWC and ER follow with 6.45\% and 7.14\% respectively, and A-GEM (8.36\%) shows nearly no improvement over the FT baseline (8.37\%).

\begin{table}[ht]
	\centering
	\caption{Marker-F1 ($\uparrow$) on the combination of all three datasets after sequential training on Pitt and Delaware, initialized on the selected SME checkpoint.}
	\label{tab:f1-exp2}
	\renewcommand{\arraystretch}{1.05}
	\begin{tabular}{l r r r r}
		\toprule
		                  & \tblscore{FILLER} & \tblscore{REP} & \tblscore{DISRUPT} & \tblscore{PAUSE} \\
		\midrule
        \tblmethod{Joint} & 0.71\pms{0.01} & 0.54\pms{0.02} & 0.42\pms{0.02} & 0.25\pms{0.02} \\
		\tblmethod{FT}    & 0.69\pms{0.02}     & 0.42\pms{0.02}  & 0.34\pms{0.00}      & 0.09\pms{0.06}    \\
		\midrule
		\tblmethod{A-GEM} & 0.68\pms{0.01}     & 0.35\pms{0.03}  & 0.34\pms{0.03}      & 0.05\pms{0.02}    \\
		\tblmethod{ER}    & \textbf{0.75}\pms{0.01} & 0.61\pms{0.00}  & \textbf{0.38}\pms{0.01} & \textbf{0.23}\pms{0.02} \\
		\tblmethod{EWC}   & 0.72\pms{0.01}     & 0.49\pms{0.02}  & 0.37\pms{0.02}      & 0.16\pms{0.01}    \\
		\tblmethod{WA}    & 0.73\pms{0.00}     & \textbf{0.63}\pms{0.01} & 0.37\pms{0.01} & 0.09\pms{0.01}    \\
		\bottomrule
	\end{tabular}
\end{table}

\section{Conclusion}

We compared several continual learning (CL) methods to investigate their ability to integrate disfluency tokens for more verbatim transcriptions and better performance on disfluent speech. We specifically focused on realistic model-expansion scenarios without joint retraining on large disfluency datasets. Overall, CL improved sequential training performance for both ASR performance and disfluency-marker retention, but we found that the best CL method depended on the primary objective. We evaluated performance in terms of initial marker inclusion, marker retention, and ASR performance on both disfluent and non-disfluent speech.

We found that CL strategies are not equally effective at introducing disfluency markers while preserving ASR quality. Across CL settings, marker F1 and pWER showed a consistent trade-off: methods that maintained the strongest pWER performance tended to under-introduce markers, while those that reliably emitted markers did so at a measurable ASR cost.
Further investigation of the effect showed that successful marker learning was linked to a consistent cross-attention pattern: a small set of heads was consistently active during marker production, and we found that targeted ablations reduced marker emission while largely preserving pWER.
This points to a reliable learned attention mechanism and offers a mechanistic explanation for why some CL methods struggle more with marker introduction.

When extending a model with additional disfluency datasets, ER achieved the strongest marker generalization and retention, with the best overall marker performance and minimal forgetting.
For the preservation of disfluent ASR performance, WA was the most stable across incremental updates, consistently delivering the strongest pWER retention and the lowest forgetting. This stability also transferred to clean speech, where WA performed best in both experiments.

While we limited ourselves to a single backbone model and a single realistic task ordering, overall, this work presented one of the first studies of CL for disfluency modeling, offering both practical guidance for verbatim ASR adaptation and mechanistic insights into how disfluency token behaviors are acquired in pretrained models. These insights provide a foundation for future work to combine and design methods that enable robust domain adaptation and reliable marker retention.

\section{Acknowledgements}
The authors gratefully acknowledge funding from Horizon Europe under the MSCA grant agreements No 101072488 (TRAIL), No 101226624 (GREET), and No 101168792 (SWEET), as well as from the German Research Foundation (DFG), project number 551629603.

Data collection for the Delaware Corpus was supported by the National Institute of Aging of the National Institutes of Health under award number RF1AG083823 (PIs: MacWhinney and Lanzi).
The Pitt Corpus is part of DementiaBank, which is funded by NIH grants NIA AG03705 and AG05133.

\section{Generative AI Use Disclosure}
Generative AI tools have been used to polish the language in the manuscript.

\bibliographystyle{IEEEtran}
\bibliography{mybib}

\end{document}